\documentclass[sigconf,natbib=false]{acmart}

\AtBeginDocument{%
  \providecommand\BibTeX{{%
    Bib\TeX}}}

\acmYear{2023}
\RequirePackage[
  datamodel=acmdatamodel,
  style=acmnumeric,
  ]{biblatex}

\newcommand{\up}[1]{{\color{red}$\uparrow#1$}}
\newcommand{\down}[1]{{\color{gray}$\downarrow#1$}}

\usepackage{times}
\usepackage{soul}
\usepackage{url}
\usepackage[utf8]{inputenc}
\usepackage{graphicx}
\usepackage{amsmath}
\usepackage{amsthm}
\usepackage{booktabs}
\usepackage{multirow}
\usepackage{algorithm}
\usepackage{algorithmic}
\usepackage[switch]{lineno}
\usepackage[capitalize]{cleveref}
\usepackage{pifont}
\usepackage{bm}
\usepackage{colortbl}  
\usepackage{xcolor}
\usepackage{array}   

\begin{document}

\title{Lightweight Super-Resolution Head for Human Pose Estimation}

\author{Haonan Wang}
\affiliation{%
  \institution{State Key Laboratory for Novel Software Technology,\\
  Nanjing University,}
  \city{Nanjing}
  \country{China}
}
\email{wanghaonan0522@gmail.com}

\author{Jie Liu}
\authornote{Corresponding author}
\affiliation{%
  \institution{State Key Laboratory for Novel Software Technology,\\
  Nanjing University,}
  \city{Nanjing}
  \country{China}
}
\email{liujie@nju.edu.cn}

\author{Jie Tang}
\affiliation{%
  \institution{State Key Laboratory for Novel Software Technology,\\
  Nanjing University,}
  \city{Nanjing}
  \country{China}
}
\email{tangjie@nju.edu.cn}

\author{Gangshan Wu}
\affiliation{%
  \institution{State Key Laboratory for Novel Software Technology,\\
  Nanjing University,}
  \city{Nanjing}
  \country{China}
}
\email{gswu@nju.edu.cn}

\renewcommand{\shortauthors}{Wang et al.}

\begin{abstract}
  Heatmap-based methods have become the mainstream method for pose estimation due to their superior performance. However, heatmap-based approaches suffer from significant quantization errors with downscale heatmaps, which result in limited performance and the detrimental effects of intermediate supervision. Previous heatmap-based methods relied heavily on additional post-processing to mitigate quantization errors. Some heatmap-based approaches improve the resolution of feature maps by using multiple costly upsampling layers to improve localization precision. To solve the above issues, we creatively view the backbone network as a degradation process and thus reformulate the heatmap prediction as a Super-Resolution (SR) task. We first propose the SR head, which predicts heatmaps with a spatial resolution higher than the input feature maps (or even consistent with the input image) by super-resolution, to effectively reduce the quantization error and the dependence on further post-processing. Besides, we propose SRPose
  to gradually recover the HR heatmaps from LR heatmaps and degraded features in a coarse-to-fine manner. To reduce the training difficulty of HR heatmaps, SRPose applies SR heads to supervise the intermediate features in each stage. In addition, the SR head is a lightweight and generic head that applies to top-down and bottom-up methods. Extensive experiments on the COCO, MPII, and CrowdPose datasets show that SRPose outperforms the corresponding heatmap-based approaches. 
  The code and models are available at \url{https://github.com/haonanwang0522/SRPose}.
\end{abstract}


\received{20 February 2007}
\received[revised]{12 March 2009}
\received[accepted]{5 June 2009}

\maketitle

\section{Introduction}

2D human pose estimation (HPE) is one of the fundamental tasks in computer vision~\cite{chen20222d}. Its purpose is to localize all the human anatomy keypoints from a single image. Since it is the basis for many human-centric visual understanding tasks such as 3D human pose estimation~\cite{zeng2021learning,zou2021eventhpe,garau2021deca,wehrbein2021probabilistic}, human action recognition~\cite{wang2013action,ji20123d,baccouche2011sequential} and pose tracking~\cite{girdhar2018detect,wang2020combining,xiao2018simple}, it has attracted widespread attention from academia and industry.

\begin{figure}[t] 
\centering
\includegraphics[scale=0.4]{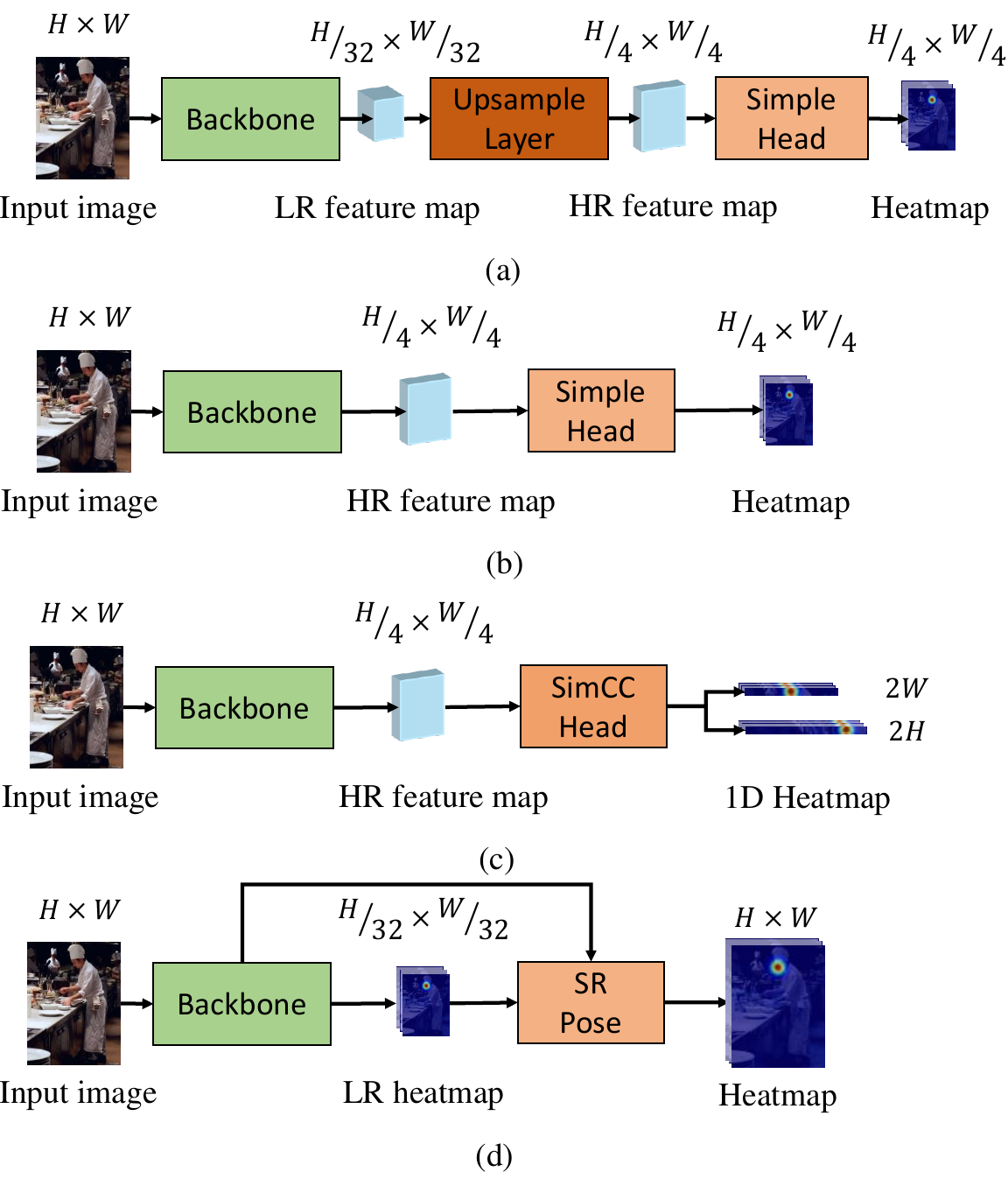}
\vspace{-0.1in}
\caption{Comparison of SRPose with previous heatmap-based methods. (a) shows the method that upsamples low-resolution feature maps by heavily upsampling layers. (b) shows the method that learns high-resolution feature maps directly. (c) shows the method that learns horizontal 1D heatmaps and vertical 1D heatmaps. (d) shows our method to obtain heatmaps with $\bm{H\times W}$ resolution from heatmaps with $\bm{\frac{H}{32}\times \frac{W}{32}}$ resolution by SR Head.}
\label{fig:intro1}
\vspace{-0.3in}
\end{figure}

2D HPE can be broadly classified into two frameworks: top-down~\cite{sun2019deep,xiao2018simple} and bottom-up~\cite{newell2017associative,cheng2020higherhrnet}. Top-down methods first detect human instances by detectors and then locate the keypoints of each human instance. However, bottom-up methods pinpoint human keypoints at first and then assign them to different human instances. In recent years, heatmap-based methods have achieved superior performance in both, especially in the top-down framework, which has become the \emph{de facto} standard. Specifically, the heatmap-based methods generate a heatmap for each kind of keypoints. Each heatmap contains a 2D Gaussian distribution centered at the ground-truth joint position, suppressing false positives and smoothing the training process. As a result, the heatmap-based methods have a more robust generalization and are easier to optimize than those that directly regress coordinates. 

Though achieving superior performance, the heatmap-based methods suffer from non-negligible quantization errors since continuous coordinate values need to be discretized into downscale heatmaps, and the resolution of the heatmaps limits the coordinate precision. The presence of quantization errors makes numerous heatmap-based methods achieve poor performance when decoding directly. 
Most heatmap-based methods rely heavily on further post-processing (\textit{e.g.}, empirical shifts, DARK~\cite{zhang2020distribution}) to mitigate quantization errors, but such post-processing is poorly optimized. In addition, some methods alleviate quantization errors by increasing the resolution of the heatmaps, and we summarize three typical ways in \cref{fig:intro1}. As shown in \cref{fig:intro1}(a), \cite{xiao2018simple,xu2022vitpose} extract Low-Resolution (LR) features (\textit{e.g.}, $\frac{H}{32}\times \frac{W}{32}$)  with rich semantic information via the backbone network. The resolution is then increased to $\frac{H}{4}\times \frac{W}{4}$ by upsampling layers. Finally, the High-Resolution (HR) features are fed into a simple prediction head (usually a $1\times 1$ convolution) to obtain heatmaps with $\frac{H}{4}\times \frac{W}{4}$ resolution.

However, the computational overhead of the upsampling layers is usually huge (\textit{e.g.}, costly deconvolution layer in \cite{xiao2018simple,xu2022vitpose}) to achieve better performance. 
As shown in \cref{fig:intro1}(b), \cite{sun2019deep,yuan2021hrformer} directly extract HR features from the input images and feed them into a simple prediction head to yield keypoint heatmaps. However, to prevent excessive computational overhead, the finest resolution of the intermediate feature is only $\frac{H}{4}\times\frac{W}{4}$. When mapping back to the original image, there is still a large quantization error. This paper aims to mitigate quantization errors with higher resolution without excessive computational overheads. 
Recently, the SimCC~\cite{li2022simcc} approach uniformly divides each pixel into several bins and reformulates HPE as two classification tasks for horizontal and vertical coordinates, as shown in \cref{fig:intro1}(c). In this way, SimCC achieves sub-pixel localization precision and low quantization error. Nevertheless, SimCC is hard to be applied to bottom-up methods directly because there may be multiple instances for each kind of joint.

\begin{figure}[t] 
\centering
\includegraphics[scale=0.4]{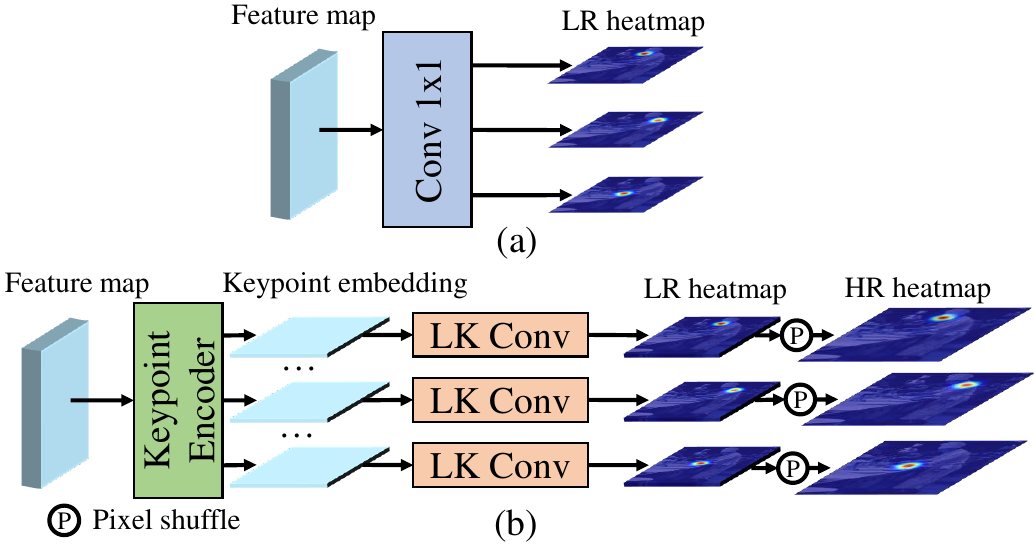}
\vspace{-0.1in}
\caption{Comparison of SR head and Simple head. (a) illustrates the simple head. (b) illustrates the SR head, where `LK Conv' indicates large kernel convolution.}
\label{fig:intro2}
\vspace{-0.3in}
\end{figure}

In this paper, we creatively view the backbone network as a degradation process and thus reformulate the heatmap prediction as a Super-Resolution (SR) task. So we can learn from the successful experience in the SR task to design a lightweight SR head for HR heatmap generation.
Specifically, as shown in \cref{fig:intro2}(b), the SR head first encodes all kinds of keypoints to keypoint embeddings by the keypoint encoder. For the $i^{th}$ keypoint embedding $K_i$, it is decoded by separate large kernel convolution to compute multiple heatmaps for the $i^{th}$ keypoint, and finally stitch all the $i^{th}$ keypoint's heatmaps together by pixel shuffle~\cite{shi2016real} to obtain the HR heatmap for the $i^{th}$ keypoint. 
Further, although the backbone network is a degradation process, the semantic information becomes richer. Therefore, we propose SRPose to predict LR heatmaps first from LR feature maps and then use the LR heatmaps and degraded features to recover HR heatmaps from coarse to fine (see \cref{fig:intro1}(d)). Besides, during training, the application of SR heads can reduce the difficulty of obtaining high-resolution heatmaps. During inference, we only keep the last SR head to improve efficiency.

Compared with previous works, our design has three merits. (1) SR head is simple, lightweight, and suitable for bottom-up and top-down frameworks. (2) SR head can be employed to predict high-resolution heatmaps (the resolution is consistent with the input), in this way to reduce quantization errors effectively. (3) SR head can be deployed for high-resolution supervision, thus alleviating the problem that low-resolution features are difficult to supervise.

The contributions are summarized as follows:
\begin{itemize}
    \item We propose a lightweight head, called SR head, that can predict high-resolution heatmaps to improve the spatial resolution of heatmaps to reduce quantization errors. Besides, As a generic regression head, the SR head can be applied to top-down and bottom-up methods.
    \item We propose SRPose, which treats the backbone network as a degradation process to reformulate the pose estimation task as a super-resolution task. To reduce the learning difficulty of the HR heatmap, we recover the precise HR heatmaps from the degraded features and LR heatmaps in a coarse-to-fine manner and apply SR heads for supervision.
    \item We conduct comprehensive experiments on COCO, MPII, and CrowdPose datasets to verify the effectiveness of the SR head and SRPose with different backbones. 
\end{itemize}

\section{Related Work}

\noindent\textbf{Top-down method.} 
The top-down method involves detecting bounding boxes by object detectors~\cite{ren2015faster,cheng2018revisiting} and estimating keypoints for each human instance. The top-down approach can be classified into regression-based~\cite{toshev2014deeppose,carreira2016human,sun2017compositional,nie2019single,mao2022poseur} and heatmap-based 
methods~\cite{newell2016stacked,chen2018cascaded,xiao2018simple,huang2020devil}. Heatmap-based methods estimate the likelihood of each keypoint for each pixel, which has better performance and is currently dominant. 
\cite{cai2020learning} and \cite{li2019rethinking} perform multi-stage feature extraction on the input and supervise the intermediate features. The intermediate supervision is performed on the interpolated low-resolution heatmaps. \cite{xiao2018simple} proposes a simple baseline with three deconvolutions after the backbone network to predict the heatmap. \cite{sun2019deep} proposes to maintain high-resolution representations throughout the process. \cite{yuan2021hrformer} introduces Transformer~\cite{vaswani2017attention} on HRNet to incorporate the self-attentive mechanism. However, quantization error has been a significant problem for heatmap-based methods.

\begin{figure*}[t]
    \begin{center}
        \includegraphics[scale=0.65]{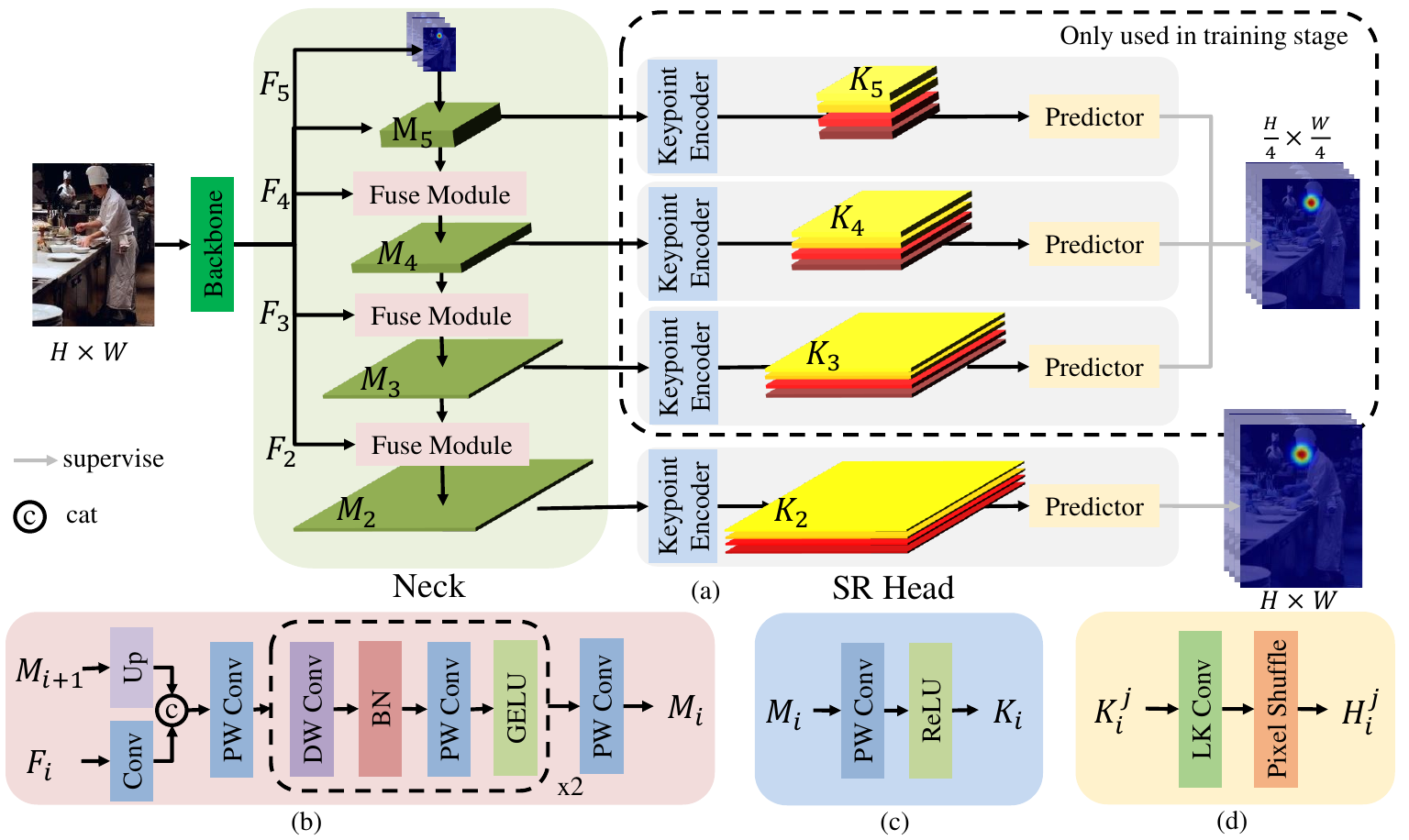}
    \end{center}
        \vspace{-0.1in}
    \caption{(a) Overview of SRPose architecture. (b) The fuse module. (c) The keypoint encoder. (d) The predictor. 
    In SRPose, the backbone is viewed as an image degradation model, and we first generate an LR heatmap by a CNN-based or Transformer-based backbone. We employed SR heads in a coarse-to-fine manner to generate HR heatmaps effectively during the training process. During inference, only the last SR head is kept for highly efficient.
    }
    \label{fig:overall}
    \vspace{-0.1in}
\end{figure*}

\noindent\textbf{Bottom-up method.} The bottom-up methods~\cite{newell2017associative,cheng2020higherhrnet} first detect all unidentified body joints in the input and then group them. \cite{cao2019openpose} learns a 2D vector field, called partial affinity field, connecting two keypoints and groups keypoints based on the integration of the line between the two keypoints. \cite{papandreou2018personlab} learns the 2D offset field of each pair of keypoints to group the keypoints. \cite{newell2017associative} proposes associative embedding for grouping, which assigns a tag to each keypoint and groups the keypoints according to the l2 distance between the tag vectors. Following it, \cite{cheng2020higherhrnet} learns a higher resolution feature pyramid based on HRNet to improve the precision of small people. 

\noindent\textbf{Quantization error.} Heatmap-based methods have seriously problematic quantization errors due to discretization and low resolution. To alleviate it, \cite{zhang2020distribution} uses a post-processing approach to synthesize the distribution information of heatmap activation by Taylor expansion-based distribution approximation.  \cite{li2022simcc} treats the pose estimation task as a classification task and tries to predict two decoupled high-resolution 1D heatmaps to improve the localization precision. However, the head of SimCC is MLP which still requires heavy weights. Besides, it applies exclusively to top-down methods, while it is difficult to apply to bottom-up methods because a heatmap contains more than one keypoint.

\noindent\textbf{Super-resolution in pose estimation.} There have been previous works~\cite{zhang2021estimating,malakshan2023joint} combining pose estimation tasks with super-resolution tasks. However, they recover the LR input through a super-resolution network to obtain an HR human image. The HR human image is later fed into the pose estimation network to estimate heatmaps. As a result, most of their inputs are of low resolution, as a higher resolution input would entail expensive computational overhead. Therefore, it is not suitable for mitigating quantization errors with this approach. Our method first proposes the SR head, which enables the model to learn HR heatmaps from LR feature maps. Besides, we treat the pose estimation task as a super-resolution task. Therefore, we recover the HR heatmaps from LR heatmaps and features in a coarse-to-fine manner. Meanwhile, we perform stage-by-stage HR supervision through SR heads for more precise recovery.

\section{Method}

The key idea of our method is to generate HR heatmaps from LR heatmaps and features to alleviate quantization errors. As shown in \cref{fig:overall}(a), we propose a general training and inference framework dubbed SRPose for more accurate HPE.
SRPose mainly consists of a backbone for feature extraction and LR heatmap generation, a neck for hierarchical feature fusion, and multiple SR heads for HR heatmap prediction and supervision. 

\subsection{Architecture of SRPose}
\label{sec:Architecture}


\noindent\textbf{Backbone.} SRPose can be applied to both Transformer-based and CNN-based backbones. In this paper, we creatively view the backbone as an image degradation model. For an input image $\bm{I}\in\mathbb{R}^{H\times W\times 3}$, the backbone can extract features and generate LR heatmaps as follows:
\begin{equation}\label{eq:degradation}
\bm{F}_{2},\bm{F}_{3},\bm{F}_{4},\bm{F}_{5},\bm{M}_{LR} = \mathcal{D}(\bm{I}),
\end{equation}
where $\mathcal{D}$ denotes the degradation process through the backbone, $\bm{F}_{k}$ represents the feature which has strides of $2^k$ pixels with respect to the input image and $\bm{M}_{LR}$ is the LR heatmaps.
After obtaining the LR heatmaps, we design a novel SRPose to generate HR heatmaps effectively.
The core part of the proposed SRPose is the SR head which will be described in detail next.

\noindent\textbf{SR Head.} LR heatmaps lead to significant quantization errors. During decoding, the precision of HPE heavily depends on the resolution of heatmaps. In existing HPE networks,  the resolution of heatmaps has already reached $\frac{H}{4}\times\frac{W}{4}$, while the quantization error is still unacceptable.  
Except for the final heatmaps, the supervision of intermediate features also suffers from quantization errors. 
For supervision, \cite{cai2020learning,li2019rethinking} generate LR heatmaps from LR intermediate features. The resolution of heatmaps is extended to $\frac{H}{4}\times\frac{W}{4}$ by corner-aligned interpolation. Although the resolution is increased by interpolation, the precision is still limited by the original heatmap. To alleviate the above problems, we propose to generate multiple heatmaps for each keypoint, and then use a pixel-shuffle~\cite{shi2016real} layer to combine multiple LR heatmaps into an HR heatmap. Specifically, as shown in \cref{fig:overall}(c), we extract several embeddings for each keypoint from the feature by the keypoint encoder as:
\begin{align}\label{eq:encoder}
\bm{K}^{0}\cdots\bm{K}^{N-1} = \text{Encoder}(\bm{M}),
\end{align}
where $\bm{M}$ is the feature to be decoded, and $\bm{K}$ denotes the 
embedding of the keypoint. There are $N$ keypoints in total.
Then, each keypoint feeds its embedding into the corresponding Large Kernel Convolution (LKC) layer to generate heatmaps as:
\begin{align}
\bm{h}^{j} = \text{LKC}^{j}(\bm{K}^{j}),
\end{align}
\begin{figure}[t]
    \begin{center}
        \includegraphics[scale=0.45]{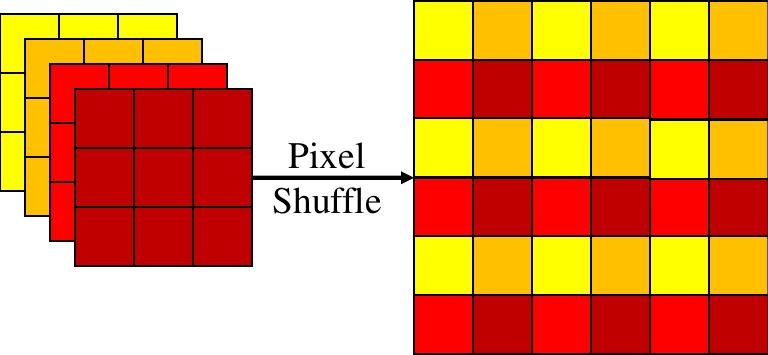}
    \end{center}
    \caption{Schematic diagram of the Pixel Shuffle layer.}
    \label{fig:pixelshuffle}
    \vspace{-0.2in}
\end{figure}
where $\bm{h}^{j}$  donates the LR heatmaps for the $j$-th keypoint. $\bm{h}^{j}$ contains $C$ heatmaps and the number $C$ is
determined by the upsampling factor of the heatmaps. Specifically, if an HR heatmap with $(l\cdot h)\times (l\cdot w)$ resolution is obtained from a feature map with $h\times w$ resolution, $C=l^2$ LR heatmaps need to be generated for each keypoint. $\text{LKC}^{j}$ denotes the convolution layer for the $j$-th keypoint. We set large kernels in LKC to better discriminate keypoints to capture larger perceptual fields. And then, we combine LR heatmaps to form an HR heatmap via pixel shuffle as follows:
\begin{align}
\bm{H}^{j} = \text{PixelShuffle}(\bm{h}^{j},l),
\end{align}
where $\bm{H}^{j}$ denotes the HR heatmap for the $j$-th keypoint and $l$ denotes the upsampling factor of pixel shuffle. The details of the Pixel Shuffle layer are shown in the \cref{fig:pixelshuffle}. In this way, we can convert the information in the channels to spatial resolution. As a result, we can obtain HR heatmaps without HR feature maps. Therefore, we do not have to extract HR features by deconvolution to reduce the parameters and computational overhead effectively. In addition, each HR heatmap is generated from its keypoint embedding, which makes the SR head more lightweight even with large kernel convolution due to the small number of channels in the keypoint embedding.

\noindent\textbf{HR Supervision.} The proposed SR head needs to increase the LR heatmaps (\textit{e.g.}, $\frac{H}{32}\times\frac{W}{32}$) to the final HR heatmaps (\textit{e.g.}, $H\times W$). However, optimizing the network at such a significant scale is extremely difficult. To alleviate the training 
difficulty, we propose to utilize multiple SR heads in the SRPose. As shown in \cref{fig:overall}(a), the SRPose
contains four SR heads, among which the first three heads are responsible for HR supervision, and the last head is designated for final HR heatmaps. In other words, the first three SR heads are only used in the training stage and will be removed during inference. As we have discussed, current supervision methods actually supervise the intermediate feature at a relatively low resolution. Different from these methods, our SR head achieves HR supervision at a spatial resolution of $\frac{H}{4}\times\frac{W}{4}$.

As shown in \cref{fig:overall}(a), we adopt an FPN-like~\cite{lin2017feature} neck to generate input
features for each SR head. Specifically, we feed the extracted feature maps from the backbone into feature fusion modules to form more representative features for SR heads:
\begin{align}
\bm{M}_{i} = \text{Fusion}(\bm{F}_i,\bm{M}_{i+1}),
\end{align}
where $i$ iterates from 5 to 2, and $M_6$ is the LR heatmaps (\textit{i.e.}, $M_{LR}$ in \cref{eq:degradation}). $\bm{M}_{i}$ serves as the input for the corresponding SR head, which is formally described in \cref{eq:encoder}. When $i=5$, Fusion(.) denotes convolution, and when $i\not=5$, Fusion(.) denotes the ``Fuse Module'' which is shown in \cref{fig:overall}(b).

\noindent\textbf{Bottom up.} The bottom-up approaches not only locate keypoints but also group keypoints to each instance, so we only replace the simple head in the baseline with the SR head to generate HR heatmaps. We use the simple method of associative embedding to perform the grouping. Associate embeddings are generated from the feature with $\frac{H}{4}\times\frac{W}{4}$ resolution by $1\times 1$ convolution. To unify the resolution, we interpolate the associate embeddings:
\begin{align}
\begin{split}
\bm{ae} &= \text{Conv}_{1\times 1}(\bm{F}_{2}),\\
\bm{AE} &= \text{Interpolate}(\bm{ae},l),
\end{split}
\end{align}
where $\bm{ae}$ donates the LR associate embeddings, $\bm{AE}$ donates the HR associate embeddings, and $l$ denotes the scale factor between the LR associate embeddings and the HR heatmaps.

\subsection{Training Targets and Loss Functions}
\label{sec:Supervision}

Heatmaps indicate the coordinates of keypoints with spatial confidence. It is generally designed to follow an unnormalized Gaussian distribution:
\begin{align}
 G(\bm{c};\bm{\mu},\sigma) = \text{exp}(-\frac{(\bm{c}-\bm{\mu})^{\text{T}}(\bm{c}-\bm{\mu})}{2\sigma^2}),
\end{align}
where $\bm{c}$ denotes the coordinates of the heatmap pixel $(c_x,c_y)$, $\bm{\mu}$ denotes the coordinates of the target joint $(\mu_x,\mu_y)$, $\sigma$ denotes the pre-defined constant. To make the before-and-after supervision consistent, we set $\sigma$ as linear to the resolution. If $\sigma=2$ for heatmaps with resolution $\frac{H}{4}\times \frac{W}{4}$, then $\sigma=8$ for heatmaps with resolution $H\times W$. For all heatmaps, we regard the mean square error (MSE) as the loss function for supervision.

\subsection{Inference}
\label{sec:Inference}

\begin{figure}[t]
    \begin{center}
        \includegraphics[scale=0.5]{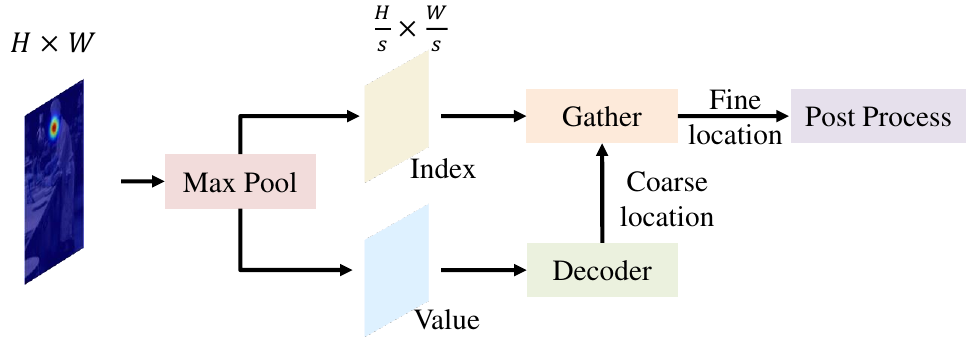}
    \end{center}
    \caption{Overview of the our inference process, which reduces inference time without compromising precision via max pooling.}
    \label{fig:inference}
    \vspace{-0.1in}
\end{figure}


We generate heatmaps with higher resolution, which effectively alleviates the problem of quantization errors, but also increases inference time. Because of the finer granularity of the candidate regions, we have to find the global maximum from more candidate regions. To alleviate the problem of our long inference time, we optimize the inference process of the heatmap. Coordinates decoding in previous heatmap-based methods is computed serially on the CPU, leading to considerable inference latency. We design a new decoding paradigm to reduce the long decoding time without influencing inference accuracy.

Briefly, we adopt the idea of divide-and-conquer, first dividing the image into multiple non-overlap patches, and then parallelizing the computation in each patch by GPU to get the maximum value of each patch and the location of the maximum value, and then finding the maximum value among all the great values. The original location of the maximum value is the final location. Specifically, as shown in \cref{fig:inference}, we utilize max pooling to reduce the resolution of heatmaps. 
\begin{align}
 \bm{Max}_{v}, \bm{Max}_{i} = \text{MaxPooling}_s(\bm{H}_2),
\end{align}
where $\bm{Max}_{v}$ and $\bm{Max}_{i}$ denote the maximum value and corresponding index of each patch. $\text{MaxPooling}_s$ denotes the max-pooling layer whose kernel size and stride are both $s$. To reduce the inference time, instead of decoding on the original heatmap, we decode $\bm{Max}_{v}$ to get the coarse location:
\begin{align}
 \bm{Loc}_{coarse} = \text{Decoder}(\bm{Max}_{v}).
\end{align}
Finally, the fine location is gathered by the index of $\bm{Max}_{i}$ according to the coarse location:
\begin{align}
 \bm{Loc}_{fine} = \bm{Max}_{v}[\bm{Loc}_{coarse}].
\end{align}
Note that the decoding process can be parallelized on GPU devices, and the decoding speed is increased without compromising accuracy. Because we reduce the spatial resolution by max pooling, and we also keep the position of the maximum value in each patch in the original HR heatmap. Finally, we can get the global maximum value and its position in the original HR heatmap, and it does not affect the precision of inference. The detailed results are shown in \cref{fig:KernelSize} of \cref{exp:ab}.

\section{Experiments}

\begin{table*}[!htb]
\centering
\caption{Comparison with different heatmap-based methods on the COCO val set. We conduct experiments on both top-down and bottom-up methods. We use the same human detection frame for a fair comparison of the top-down methods. We tested the results with (w/ Post) and without (w/o Post) adding an empirical shift for each top-down method. The size of the input for all top-down methods is $\bm{256\times 192}$, while the size of the input for bottom-up methods is $\bm{512\times 512}$. Our method provides significant gains for all backbones and reduces the dependence on refinement post-processing.}
\vspace{-0.1in}
\label{tab:CocoVal}
\scalebox{1.0}{
\begin{tabular}{lcccccccc}
\toprule
\multirow{2}{*}{\textbf{Backbone}} & \multirow{2}{*}{\textbf{Scheme}} & \multirow{2}{*}{\textbf{GFLOPs}} & \multicolumn{2}{c}{\textbf{Params}} &\multicolumn{2}{c}{\textbf{w/ Post.}} & \multicolumn{2}{c}{\textbf{w/o Post.}}\\
\cmidrule(lr){4-5} \cmidrule(lr){6-7} \cmidrule(lr){8-9} &  &  & \textbf{Backbone} & \textbf{Other} & \textbf{AP} & \textbf{AR} & \textbf{AP} & \textbf{AR} \\ \midrule
\multicolumn{9}{c}{\textbf{Top-down methods}}\\ \midrule

\multirow{4}{*}{Resnet-50~\cite{he2016deep}} 
 & Simple head    & 5.46  &23.51M(69\%) &10.49M(31\%) & 71.7(\down{0.0}) & 77.3(\down{0.0}) & 69.8(\down{1.9}) & 75.8(\down{1.5}) \\ \cmidrule{2-9} 
 & SR head        & 5.77  &23.51M(69\%) &10.59M(31\%) & 72.4(\up{0.7}) & 77.9(\up{0.6}) & 72.2(\up{0.5}) & 77.7(\up{0.4}) \\ \cmidrule{2-9} 
 &\textbf{SRPose} & 4.61  &23.51M(95\%) &1.29M(5\%)   & \textbf{73.3}(\up{1.6}) & \textbf{78.8}(\up{1.5}) & 73.1(\up{1.4}) & 78.6(\up{1.3}) \\ \midrule
 
\multirow{4}{*}{HRNet-W32~\cite{sun2019deep}} 
 & Simple head    & 7.70  &28.54M(100\%) & 0.00M(0\%) & 74.5(\down{0.0}) & 79.9(\down{0.0}) & 72.3(\down{2.2}) & 78.2(\down{1.7}) \\ \cmidrule{2-9} 
 & SR head        & 7.98  &28.54M(100\%) & 0.09M(0\%) & 75.6(\up{1.1})   & 80.6(\up{0.7}) & 75.4(\up{0.9})   & 80.5(\up{0.6}) \\ \cmidrule{2-9} 
 &\textbf{SRPose} & 8.28  &29.30M(98\%) & 0.65M(2\%)  & \textbf{75.9}(\up{1.4}) & \textbf{81.0}(\up{1.1}) & 75.7(\up{1.2}) & 80.9(\up{1.0}) \\ \midrule

\multirow{4}{*}{TransPose-R-A4~\cite{yang2021transpose}} 
 & Simple head    & 8.91  &4.93M(82\%) & 1.06M(18\%) & 71.8(\down{0.0}) & 77.3(\down{0.0}) & 69.7(\down{2.1}) & 75.5(\down{1.8}) \\ \cmidrule{2-9} 
 & SR head        & 9.23  &4.93M(81\%)  &1.16M(19\%)  & 73.2(\up{1.4})   & 78.4(\up{1.1}) & 73.1(\up{1.3})   & 78.3(\up{1.0}) \\ \cmidrule{2-9} 
 &\textbf{SRPose} & 6.26  &4.93M(90\%) & 0.55M(10\%)  & \textbf{73.5}(\up{1.7}) & \textbf{78.9}(\up{1.6}) & 73.4(\up{1.6}) & 78.7(\up{1.4}) \\  \midrule
 
\multirow{4}{*}{HRFormer-S~\cite{yuan2021hrformer}} 
 & Simple head    & 2.82  &7.89M(100\%) & 0.00M(0\%) & 74.0(\down{0.0}) & 79.2(\down{0.0}) & 72.1(\down{1.9}) & 77.6(\down{1.6}) \\ \cmidrule{2-9} 
 & SR head        & 3.09  &7.89M(99\%)  &0.09M(1\%)  & 75.0(\up{1.0})   & 80.1(\up{0.9}) & 74.8(\up{0.8})   & 80.0(\up{0.8}) \\ \cmidrule{2-9} 
 &\textbf{SRPose} & 3.34  &8.21M(93\%) & 0.65M(7\%)  & \textbf{75.6}(\up{1.6}) & \textbf{80.7}(\up{1.5}) & 75.5(\up{1.5}) & 80.6(\up{1.4}) \\  \midrule
\multicolumn{9}{c}{\textbf{Bottom-up methods}}\\ \midrule

\multirow{2}{*}{Resnet-50~\cite{he2016deep}} 
 & Simple head     & 29.20 &23.51M(69\%) &10.49M(31\%) & 46.7(\down{0.0}) & 55.1(\down{0.0}) & - & -   \\ \cmidrule{2-9} 
 & \textbf{SR head} & 30.86 &23.51M(69\%) &10.60M(31\%) & \textbf{48.4}(\up{1.7}) & \textbf{56.6}(\up{1.5}) & - & -  \\  \midrule
 
\multirow{2}{*}{HRNet-W32~\cite{sun2019deep}} 
 & Simple head     & 41.10 &28.54M(100\%) &0.00M(0\%) & 65.3(\down{0.0}) & 70.9(\down{0.0}) & - & - \\ \cmidrule{2-9} 
 &\textbf{SR head} & 42.57 &28.54M(100\%) & 0.09M(0\%) & \textbf{67.1}(\up{1.8}) & \textbf{71.7}(\up{0.8}) & - & - \\
 
 \bottomrule
\end{tabular}

}
\vspace{-0.1in}
\end{table*}

In the following sections, we conduct experiments on several datasets to validate the effectiveness of SRPose and SR head for human pose estimation. Experiments are conducted on three benchmark datasets: COCO~\cite{lin2014microsoft}, MPII~\cite{andriluka20142d}, and CrowdPose~\cite{li2019crowdpose}. The ablation experiments are conducted on the benchmark dataset of COCO.

\subsection{COCO Keypoint Detection}
\label{exp:coco}

\noindent\textbf{Dataset.} The COCO dataset~\cite{lin2014microsoft} contains more than $200,000$ images and $250,000$ human instances which are labeled with 17 keypoints. It is divided into three sets, the train set is with $57k$ images, the val set is with $5k$ images, and the test-dev set is with $20k$ images. Experimental results are reported on both the test-dev set and the val set. The data augmentation settings follow MMPose~\cite{mmpose2020}.

\noindent\textbf{Evaluation metric.} The standard evaluation metric for the COCO dataset is the standard average precision(AP), which is based on Object Keypoint Similarity(OKS):
\begin{equation}
    OKS=\frac{\sum_i\text{exp}(-d^2_{p^i}/2S^2_p\sigma^2_i)\delta(v_{p^i}>0)}{\sum_i\delta(v_{p^i}>0)},
\end{equation}
where $d_{p^i}$ indicates the Euclidean distance between $i^{th}$ keypoint of human $p$ and the corresponding ground truth, $v_{p^i}$ indicates visibility of the keypoint, $S_p$ indicates the scale factor of human $p$, and $\sigma_i$ indicates the factor of keypoint $i$. 

\noindent\textbf{Backbone settings.} Currently, there are many backbones for human pose estimation. For top-down methods, the backbone can be broadly classified into CNN-based and Transformer-based. To demonstrate the applicability of SRPose to both types of backbone, we choose two state-of-the-art methods (\textit{i.e.}, Resnet~\cite{he2016deep} and HRNet~\cite{sun2019deep}) from the CNN-based methods and two (i.e. TransPose~\cite{yang2021transpose} and HRFormer~\cite{yuan2021hrformer}) from the Transformer-based methods as baselines. For bottom-up methods, we take Resnet and HRNet as the backbone.

\noindent\textbf{Implementation details.} For the selected backbones, we simply follow the settings in MMPose. Specifically, for top-down methods, all the models are trained with batch size 128 (batch size 128 for HRFormer-B due to limited GPU memory) and are optimized by Adam (AdamW~\cite{loshchilov2017decoupled} for HRFormer in MMPose) with a base learning rate of $5\times10^{-4}$ decreased to  $5\times10^{-5}$ at the $170^{th}$ epoch, to $5\times10^{-6}$ at the $200^{th}$ epoch and ended at the $210^{th}$; $\beta_1$ and $\beta_2$ are set to $0.9$ and $0.999$; weight decay is set to $10^{-4}$. Meanwhile, they use the two-stage top-down human pose estimation as the pipeline, which first detects the human by a detector and then crops it out to estimate its keypoints. Following~\cite{xiao2018simple}, we adopt the person detector with $56.4\%$ for COCO val set. For bottom-up methods, all the models are trained with batch size 48 and are optimized by Adam with a base learning rate of $1.5\times10^{-3}$ dropped to  $1.5\times10^{-4}$ at the $200^{th}$ epoch, to $1.5\times10^{-5}$ at the $260^{th}$ epoch and ended at the $300^{th}$. 

\noindent\textbf{Results on the COCO val set.}
We conduct extensive experiments based on different backbones on the COCO val set to validate the effectiveness of our proposed SR head and SRPose framework. For top-down methods, some well-performing CNN-based and Transformer-based methods are selected as the baseline, while the bottom-up methods only chose CNN-based methods as the baseline due to the oversized inputs. The results in \cref{tab:CocoVal} illustrate that the SR head is efficient and shows consistent performance dominance over the Simple head across different backbones and frameworks, especially for the bottom-up methods. For instance, in the bottom-up HRNet-W32~\cite{cheng2020higherhrnet}, replacing Simple head with SR head results in a gain of 1.9 AP, but the number of parameters is increased by only 0.09M. In addition, the SR head reduces the dependence of the top-down method on post-processing due to the HR heatmap. For example, when HRNet-W32~\cite{sun2019deep} serves as the backbone, SR head drops only 0.2AP without post-processing, but Simple head drops 2.2AP. Overall, whether the backbone is pyramid-based or HR-based, SRPose can bring significant benefits with reduced or small additional parameters. For example, in ResNet-50~\cite{he2016deep}, SRPose improves 1.6AP, while in HRFormer-S, SRPose improves 1.6AP.

\begin{table}[!t]
\centering
\caption{Comparison on COCO test-dev set. `T' denotes the abbreviation of Transformer. Our method achieves state-of-the-art results with significantly improved performance compared to the baseline.}
\vspace{-0.1in}
\label{tab:COCOTest}
\scalebox{0.85}{
\begin{tabular}{llcccc}\toprule
\multicolumn{1}{l}{\textbf{Method}} &\textbf{Backbone} & \textbf{Input size} & \multicolumn{1}{c}{\textbf{GFLOPs}} & \multicolumn{1}{c}{\textbf{AP}} & \multicolumn{1}{c}{\textbf{AR}} \\ \midrule

\multicolumn{6}{c}{\textbf{Top-down methods}} \\ \midrule

\multicolumn{6}{l}{\textbf{Regression-based}} \\
\multicolumn{1}{l}{CenterNet~\cite{zhou2019objects}}& Hourglass &-& - & 63.0 & - \\
\multicolumn{1}{l}{DirectPose~\cite{tian2019directpose}}& ResNet-50 &-& \multicolumn{1}{c}{-} & 62.2 & - \\
\multicolumn{1}{l}{PointSetNet~\cite{wei2020point}}&HRNet-W48 &-& \multicolumn{1}{c}{-} & 68.7 & -\\
 \multicolumn{1}{l}{Integral Pose~\cite{sun2018integral}}& ResNet-101 &256$\times$256& \multicolumn{1}{c}{11.0} & 67.8 & - \\
\multicolumn{1}{l}{TFPose~\cite{mao2021tfpose}}& ResNet-50+T &384$\times$288& \multicolumn{1}{c}{20.4} & 72.2 & -\\
\multicolumn{1}{l}{RLE~\cite{li2021human}}& HRNet-W48 &-& \multicolumn{1}{c}{-} & 75.7 & - \\ 

\multicolumn{6}{l}{\textbf{Heatmap-based}} \\
\multicolumn{1}{l}{SimBa~\cite{xiao2018simple}}& ResNet-50 & 384$\times $288 & \multicolumn{1}{c}{ 20.0} &   71.5 & 76.9\\
\multicolumn{1}{l}{SimBa~\cite{xiao2018simple}}& ResNet-152 & 384$\times$288 & \multicolumn{1}{c}{ 35.6} &  73.7 & 79.0\\
\multicolumn{1}{l}{TransPose~\cite{yang2021transpose}} &HRNet-W48+T & 256$\times$192 & \multicolumn{1}{c}{ 21.8} & 75.0 & 80.1 \\
\multicolumn{1}{l}{TokenPose~\cite{li2021tokenpose}} &L/D24& 384$\times$288 & \multicolumn{1}{c}{ 22.1} & 75.9 & 80.8 \\
\multicolumn{1}{l}{HRNet~\cite{sun2019deep}}& HRNet-W32 & 384$\times$288 & \multicolumn{1}{c}{ 16.0} &  74.9 & 80.1\\
\multicolumn{1}{l}{HRNet~\cite{sun2019deep}} &HRNet-W48& 384$\times$288 & \multicolumn{1}{c}{ 32.9} &  75.5 & 80.5\\
\multicolumn{1}{l}{SimCC~\cite{li2022simcc}}& ResNet-50 &384$\times$288& \multicolumn{1}{c}{20.2} & 72.7 & 78.0\\
\multicolumn{1}{l}{SimCC~\cite{li2022simcc}}& HRNet-W48 &384$\times$288& \multicolumn{1}{c}{32.9} & 76.0 & 81.1\\
\multicolumn{1}{l}{DARK~\cite{zhang2020distribution}}& HRNet-W48 &384$\times$288& \multicolumn{1}{c}{32.9} & 76.2 & 81.1\\
\multicolumn{1}{l}{HRFormer~\cite{yuan2021hrformer}}& HRFormer-S &384$\times$288& \multicolumn{1}{c}{6.2} & 74.5 & 79.8\\
\multicolumn{1}{l}{HRFormer~\cite{yuan2021hrformer}}& HRFormer-B &384$\times$288& \multicolumn{1}{c}{26.8} & 76.2 & 81.2\\
\rowcolor{gray!20}\multicolumn{1}{l}{SRPose}& ResNet-50 & 384$\times $288 & \multicolumn{1}{c}{ 24.6 } &   74.1 & 79.1\\
\rowcolor{gray!20}\multicolumn{1}{l}{SRPose}& HRFormer-B & 384$\times $288 & \multicolumn{1}{c}{ 30.4 } &   \textbf{76.9} &   \textbf{81.8} \\
 \midrule
 
\multicolumn{6}{c}{\textbf{Bottom-up methods}} \\ \midrule

\multicolumn{1}{l}{OpenPose~\cite{cao2019openpose}}& - & - & - & 61.8 & 66.5\\
\multicolumn{1}{l}{AE~\cite{newell2017associative}}& Hourglass &512$\times$512& \multicolumn{1}{c}{206.9} & 65.5 & 70.2\\
\multicolumn{1}{l}{HRNet~\cite{cheng2020higherhrnet}}& HRNet-W32 &512$\times$512& \multicolumn{1}{c}{38.9} & 64.1& -\\
\multicolumn{1}{l}{HigherHRNet~\cite{cheng2020higherhrnet}}& HRNet-W32 &512$\times$512& \multicolumn{1}{c}{47.9} & 66.4& -\\
\rowcolor{gray!20}\multicolumn{1}{l}{HRNet + SR head}& HRNet-W32 &512$\times$512& \multicolumn{1}{c}{42.6} & \textbf{66.5}& \textbf{71.2}\\
\bottomrule
\end{tabular}
}
\vspace{-0.1in}
\end{table}

\noindent\textbf{Results on the COCO test-dev set.}
We perform the comparison on the COCO test-dev set, with the results shown in \cref{tab:COCOTest}. For top-down methods with $384\times 288$ input size, our method boosts 2.6AP and 0.7AP for SimpleBaseline-Res50~\cite{xiao2018simple} and HRFormer-B~\cite{yuan2021hrformer}, respectively. For bottom-up methods with $512\times 512$ input size, our method improves 2.4AP for HRNet-W32, even better than HigherHRNet~\cite{cheng2020higherhrnet} with fewer GFOLPs.

\subsection{Ablation Study}
\label{exp:ab}
    In this section, we use Resnet-50 as the backbone to conduct ablation studies for investigating our proposed model. The training settings are the same as \cref{exp:coco}, and we provide the test results of COCO val to observe the performance. The default resolution is $256\times 192$.

\begin{table}[t]
\centering
\caption{Ablation study of scale factor $\bm{k}$ on the COCO val set. $\bm{k}$ controls the resolution ratio of the output heatmap to the input image, and it improves performance at the beginning of growth but then the performance decreases.}
\vspace{-0.1in}
\label{tab:ScaleFactor}
\scalebox{1.0}{
\begin{tabular}{cccccccc}
\toprule
 \textbf{$\bm{k}$} & \textbf{Output size} & \textbf{Params} &\textbf{GFLOPs} & \textbf{AP} & \textbf{AR}\\ \midrule
$0.25$ & $64\times 48$   &24.71M & 4.36 & 72.7 & 78.2\\
$0.5$  & $128\times 96$  &24.73M & 4.41 & 73.1 & 78.6\\
$1$    & $256\times 192$ &24.80M & 4.61 & \textbf{73.3} & \textbf{78.8}\\
$2$    & $512\times 384$ &25.06M & 5.43 & 73.2 & 78.6 \\
 \bottomrule
\end{tabular}
}
\vspace{-0.1in}
\end{table}

\noindent\textbf{Scale factor $k$.} Scale factor $k$ represents the resolution ratio of the output heatmap to the input image. It is proportional to the resolution of the output heatmap and inversely proportional to quantization errors. However, it is also proportional to the number of parameters in the SR head, which may lead to overfitting easily. To balance between the model performance and quantization error, we set $k\in\{0.25,0.5,1,2\}$ to experiment. According to \cref{tab:ScaleFactor}, it is easy to find that when $k$ grows from $0.25$ to $0.5$, the model's performance has a large improvement, but as $k$ grows, the performance growth slows down. The model reaches its best when $k=1$, thus we set $k=1$ in SRPose.

\begin{table}[!t]
\centering
\caption{Ablation study of HR supervision on the COCO val set. As supervision decreases, so does performance.}
\vspace{-0.1in}
\label{tab:SuperviseNum}
\scalebox{1}{
\begin{tabular}{ccccccccc}
\toprule
 \multirow{2}{*}{\textbf{Method}} & \multicolumn{4}{c}{\textbf{Supervised Feature}} & \multirow{2}{*}{\textbf{AP}} & \multirow{2}{*}{\textbf{AR}}\\ \cmidrule{2-5}
 & $\mathbf{M_5}$ & $\mathbf{M_4}$& $\mathbf{M_3}$& $\mathbf{M_2}$& & \\ \midrule
SimBa + SR head & \ding{55} & \ding{55} & \ding{55} & \checkmark & 72.4  & 77.9\\\midrule
\multirow{4}{*}{SRPose} 
 & \checkmark & \checkmark & \checkmark & \checkmark & \textbf{73.3} & \textbf{78.8}\\
 & \ding{55} & \checkmark & \checkmark & \checkmark & 72.7 & 78.2 \\
 & \ding{55} & \ding{55} & \checkmark & \checkmark & 72.7 & 78.2\\
 & \ding{55} & \ding{55} & \ding{55} & \checkmark & 72.6 & 78.1\\
 \bottomrule
\end{tabular}
}
\vspace{-0.1in}
\end{table}
\noindent\textbf{HR supervision.} To verify the necessity of HR supervision, we conduct experiments by gradually reducing HR supervision from top to down. For better comparison, we regard SimpleBaseline + SR head as the baseline. As shown in \cref{tab:SuperviseNum}, Simply fusing the features can only bring a slight improvement to the model performance, but as the number of supervisions grows, so does the model's performance. And from the results, we can find that it is necessary to supervise the topmost feature, which can significantly improve performance. 
Because it can bring performance gains without increasing overhead in inference, we recommend supervising all features.

\begin{figure}[t]
    \begin{center}
        \includegraphics[scale=0.6]{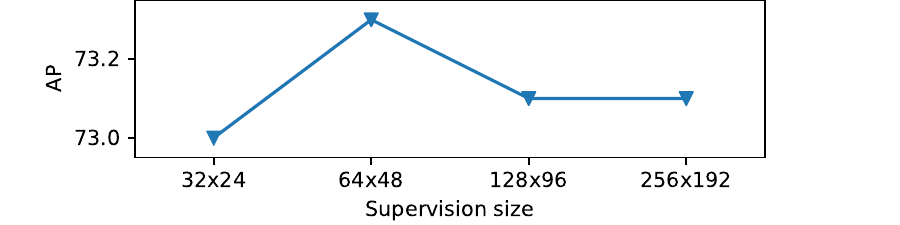}
    \end{center}
    \vspace{-0.1in}
    \caption{Ablation study of supervision size on the COCO val set. The supervision size improves performance at the beginning of growth, but then the performance decreases.}
    \label{fig:SupervisionSize}
    \vspace{-0.1in}
\end{figure}
\noindent\textbf{Resolution of HR supervision.} The higher resolution of the heatmap results in lower quantization error, which also applies to intermediate supervision. High-resolution heatmaps can help the model to supervise the merged features well. However, with the common issue of high-resolution heatmaps, there is a squared increase in the number of heatmaps, \textit{e.g.}, for a feature $y\in\mathbb{R}^{\frac{H}{32}\times\frac{W}{32}\times C}$, if we use a ground truth $G\in\mathbb{R}^{H\times W}$ to perform supervision, then 1024 heatmaps $h\in\mathbb{R}^{\frac{H}{32}\times\frac{W}{32}}$ have to be predicted for each keypoint. This not only has to learn a large number of parameters but also is not conducive to the supervision of intermediate features. To balance them, we set the resolution to $s\in\{\frac{H}{8}\times\frac{W}{8},\frac{H}{4}\times\frac{W}{4},\frac{H}{2}\times\frac{W}{2}, H\times W\}$ for our experiments. According to \cref{fig:SupervisionSize}, we can see that the performance of the model improves when the resolution grows to  $\frac{H}{4}\times\frac{W}{4}$, after which, on the contrary, it decreases. Therefore, the recommended settings is $\frac{H}{4}\times\frac{W}{4}$.

\begin{figure}[t]
    \begin{center}
        \includegraphics[scale=0.6]{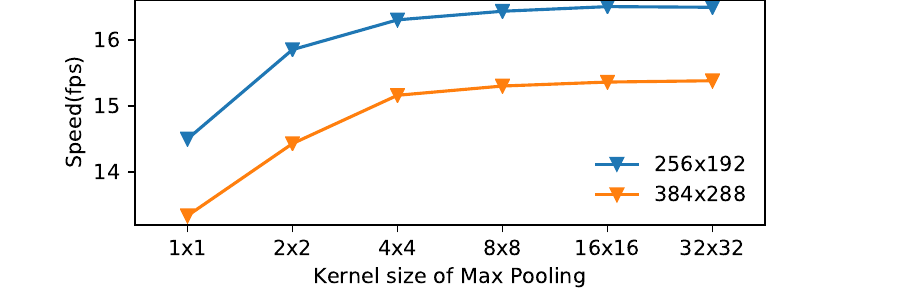}
    \end{center}
    \vspace{-0.1in}
    \caption{Ablation study of kernel size of max pooling on the COCO val set. The optimization can improve inference speed, but the speedup slows down as the kernel size increases.}
    \label{fig:KernelSize}
    \vspace{-0.2in}
\end{figure}
 \noindent\textbf{Kernel size of max pooling.}  We downsample heatmaps by max pooling to mitigate the effect of the over-resolution of heatmaps. To verify whether our optimization can improve the inference speed, we set the kernel size $ks =\{1,2,4,8,16,32\}$ and conduct experiments at two input resolutions. Note that $ks=1$ means not to use our proposed optimization but to decode the high-resolution heatmap directly. As shown in \cref{fig:KernelSize}, our proposed optimization can improve inference speed. However, as the kernel size keeps increasing, the speedup slows down, and when the kernel size reaches 16, the inference speedup almost stops. Therefore, we set the kernel size to 16 in the inference process.

\begin{table}[!t]
\centering
\caption{Ablation study of different heads on the COCO val set. We evaluate all heatmaps of the output. The SR head is significantly better than the Simple head in performance.}
\vspace{-0.1in}
\label{tab:SuperviseType}
\scalebox{1}{
\begin{tabular}{ccccc}
\toprule
 \textbf{Supervision head} & \textbf{Prediction head}& \textbf{Params} & \textbf{AP} & \textbf{AR}\\ \midrule
Inter head & Simple head & 24.71M & 69.8 & 75.7\\
SR head     & Simple head & 24.71M & 70.0 & 75.8\\
Inter head & SR head     & 24.80M & 72.8 & 78.3\\
SR head     & SR head     & 24.80M & \textbf{73.3}& \textbf{78.8} \\
 \bottomrule
\end{tabular}

}
\vspace{-0.1in}
\end{table}
\noindent\textbf{Heads for supervision and prediction.} There are previous methods (\textit{e.g.}, RSN~\cite{cai2020learning} , MSPN~\cite{li2019rethinking}) to supervise the features at different scales. However, they first generate heatmaps for each keypoint by $1\times 1$ convolution and then refine heatmaps by multiplying $3\times 3$ convolution. Finally, the heatmaps are scaled to the desired size by interpolation with corner alignment, which may result in feature misalignment. We refer to this supervision head as the ``Inter'' head. To verify the performance of the SR head in intermediate supervision and prediction, we performed with different heads as supervision and prediction heads in our experiments. The result (\cref{tab:SuperviseType}) indicates that the SR head outperforms Inter head in supervision and outperforms Simple Head in prediction.

\begin{table}[!t]
\centering
\caption{Ablation study of kernel size of SR heads on the COCO val set. As the kernel size increases, the performance is gradually improved.}
\vspace{-0.1in}
\label{tab:KernelSize}
\scalebox{1}{
\begin{tabular}{cccccccc} 
\toprule
 \multicolumn{4}{c}{\textbf{Kernel size}} & \multirow{2}{*}{\textbf{Params}}& \multirow{2}{*}{\textbf{AP}}& \multirow{2}{*}{\textbf{AR}}\\ \cmidrule{1-4} 
 $\mathbf{M_5}$ & $\mathbf{M_4}$& $\mathbf{M_3}$& $\mathbf{M_2}$& &\\ \midrule
 1 & 1 & 1 & 1 & 24.71M & 72.9 & 78.4 \\ 
 3 & 3 & 3 & 3 & 24.72M & 73.1 & 78.6 \\ 
 3 & 5 & 7 & 9 & 24.80M & \textbf{73.3} & \textbf{78.8}\\ 
 \bottomrule
\end{tabular}
}
\vspace{-0.1in}
\end{table}
\noindent\textbf{Kernel size of SR heads.} We first use the convolution of $1\times 1$ kernel to convert the feature map into keypoint embeddings. Then we use the grouped convolution intended to achieve a larger perceptual field with a smaller number of parameters. To verify the necessity of a sizeable perceptual field, we set different kernel sizes of convolution for the experiment. From \cref{tab:KernelSize}, it can be concluded that the convolution of large kernel size can achieve better performance. Therefore, we set the SR heads with increasing kernel size. Besides, as can be seen from the table, the number of parameters does not improve much as the size of the convolution kernel increases due to the small number of channels at each keypoint embedding.

\subsection{MPII Human Pose Estimation} 
\noindent\textbf{Dataset \& Evaluation metric.} We have validated the effectiveness of our proposed method on the COCO dataset, and for further validation, we conduct experiments on the MPII dataset~\cite{lin2014microsoft}. The MPII dataset contains about $25000$ images. Each image contains at least $1$ person. There are about $40000$ individuals labeled with 16 keypoints in the dataset, which is divided into $28000$ for training and the rest for testing. The head-normalized probability of correct keypoints (PCKh) is used to evaluate the performance.

\noindent\textbf{Results on the validation set.}
The results of the MPII validation set are shown in \cref{tab:mpii}. We used Res50 and HRNet-W32 as the backbone for our experiments, which improved by 0.9 and 0.5 in PCKh compared to SimpleBaseline-Res50 and HRNet-W32, respectively.

\begin{table}[!t]
\centering
\caption{Comparison with top-down methods on MPII validation set. SRPose can achieve significant performance improvement compared to baselines on different backbones. Reg: regression-based approach; HM: heatmap-based approach.}
\label{tab:mpii}
\scalebox{1.0}{
\begin{tabular}{llcc}\toprule
\multicolumn{1}{l}{\textbf{Method}} &\textbf{Backbone} & \textbf{Type} & \textbf{PCKh@0.5} \\ \midrule
RLE~\cite{li2021human}          & ResNet-50  & Reg. & 85.8\\
Integral~\cite{sun2018integral} & ResNet-101 & Reg. & 87.3\\
PRTR~\cite{li2021pose}          & HRNet-W32  & Reg. & 89.5\\
SimBa~\cite{xiao2018simple}     & ResNet-50  & HM.  & 88.2 \\
HRNet~\cite{sun2019deep}    & HRNet-W32  & HM.  & 90.1 \\
SimCC~\cite{li2022simcc}        & HRNet-W32  & HM.  & 90.0 \\
TokenPose~\cite{li2021tokenpose}& L/D24      & HM.  & 90.2 \\
\rowcolor{gray!20}SRPose        & ResNet-50  & HM.  & 89.1 \\
\rowcolor{gray!20}SRPose        & HRNet-W32  & HM.  & \textbf{90.5} \\

\bottomrule
\end{tabular}
}
\end{table}

\subsection{CrowdPose} 
\noindent\textbf{Dataset \& Evaluation metric.} To verify the performance of SRPose in dense pose scenes, we further experiment on the CrowdPose dataset~\cite{li2019crowdpose}. The  CrowdPose dataset has more crowded scenes, which is the most distinctive feature distinguished from other datasets. It contains a total of 20K images and 80K human instances. The images are divided into $10000$ for training, $2000$ for validation, and the rest for testing. In addition to the consistent evaluation metrics with COCO~\cite{lin2014microsoft}, the CrowdPose dataset has metrics $\text{AP}^{E}$ and $\text{AP}^{H}$, where $\text{AP}^{E}$ is the AP score for relatively simple ones and $AP^{H}$ is the AP score for hard ones. We follow the original paper to detect human instances by YoloV3~\cite{redmon2018yolov3} and then test performance on the test set of CrowdPose.

\begin{table}[!t]
\centering
\caption{Comparison with top-down methods on CrowdPose test set. SRPose can improve the performance of different baselines in dense pose scenes.}
\label{tab:crowdpose}
\scalebox{1.0}{
\begin{tabular}{llcccc}\toprule
\multicolumn{1}{l}{\textbf{Method}} &\textbf{Backbone} & \textbf{AP} & \textbf{${\text{AP}^E}$}& \textbf{${\text{AP}^M}$}& \textbf{${\text{AP}^H}$} \\ \midrule

SimBa~\cite{xiao2018simple}& ResNet-50 & 63.7 & 73.9 & 65.0 & 50.6 \\
HRNet-W32~\cite{sun2019deep} &HRNet-W32 & 66.4 & 74.0 & 67.4 & 55.6 \\
SimCC~\cite{li2022simcc} & HRNet-W32 & 66.7 & 74.1 & 67.8 & \textbf{56.2} \\
\rowcolor{gray!20}SRPose & ResNet-50 & 64.7 & 74.9 & 65.8 & 52.3\\
\rowcolor{gray!20}SRPose & HRNet-W32 & \textbf{67.8} & \textbf{77.5} & \textbf{69.1} & 55.6 \\

\bottomrule
\end{tabular}
}
\end{table}

\noindent\textbf{Results on the test set.}
The results of the CrowdPose test set are shown in \cref{tab:crowdpose}. From the experimental results, it can be seen that our proposed SRPose is equally effective in the scenario of the dense pose. Compared with the original SimpleBaseline-Res50 and HRNet-W32, our SRPose improves the AP by 1.0 and 1.4, respectively.

\section{Conclusion}
In this paper, we propose a brand new head, SR head, which can predict high-resolution heatmaps (the same resolution as the input) to alleviate the quantization error and prevent the dependency on post-processing with heatmap-based methods. Besides, to reduce the training difficulty for obtaining large-resolution, we propose SRPose that applies the SR head for HR supervision in a coarse-to-fine manner during the training process. Extensive experiments on COCO, MPII, and CrowdPose benchmarks demonstrate that SRPose can enhance the performance of top-down methods, and SR head can also enhance the performance of bottom-up methods.

\printbibliography
\end{document}


\title{Lightweight Super-Resolution Head for Human Pose Estimation (Technical Appendix)}

\maketitle

\section{Keypoint embedding visualization}
\begin{figure}[t]
    \begin{center}
        \includegraphics[scale=0.6]{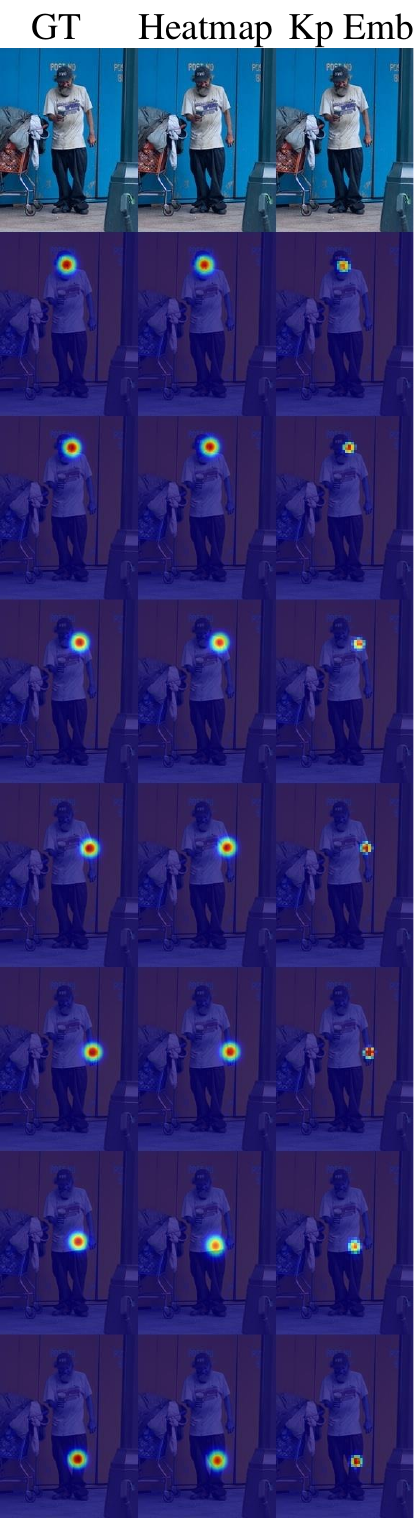}
    \end{center}
        \vspace{-0.1in}
    \caption{Keypoint embedding visualization, where `GT' denotes ground truth, `Kp Emb' denotes keypoint embedding.}
    \label{fig:KpEmb}
\end{figure}

To make an intensive study of SRPose, we set each keypoint to generate a keypoint embedding first and then visualize the keypoint embedding. As shown in \cref{fig:KpEmb}, the keypoint embedding's activation center is the same as the heatmap's. However,  the extent of the activation region is smaller than the heatmap's, so the effect of excessive sigma can be effectively mitigated. Besides, different keypoints can generate their keypoint embedding according to the characteristics of the keypoints themselves so that a higher quality HR heatmap can be obtained.

\section{Heatmap visualization}

\begin{figure*}[t]
    \begin{center}
        \includegraphics[scale=0.5]{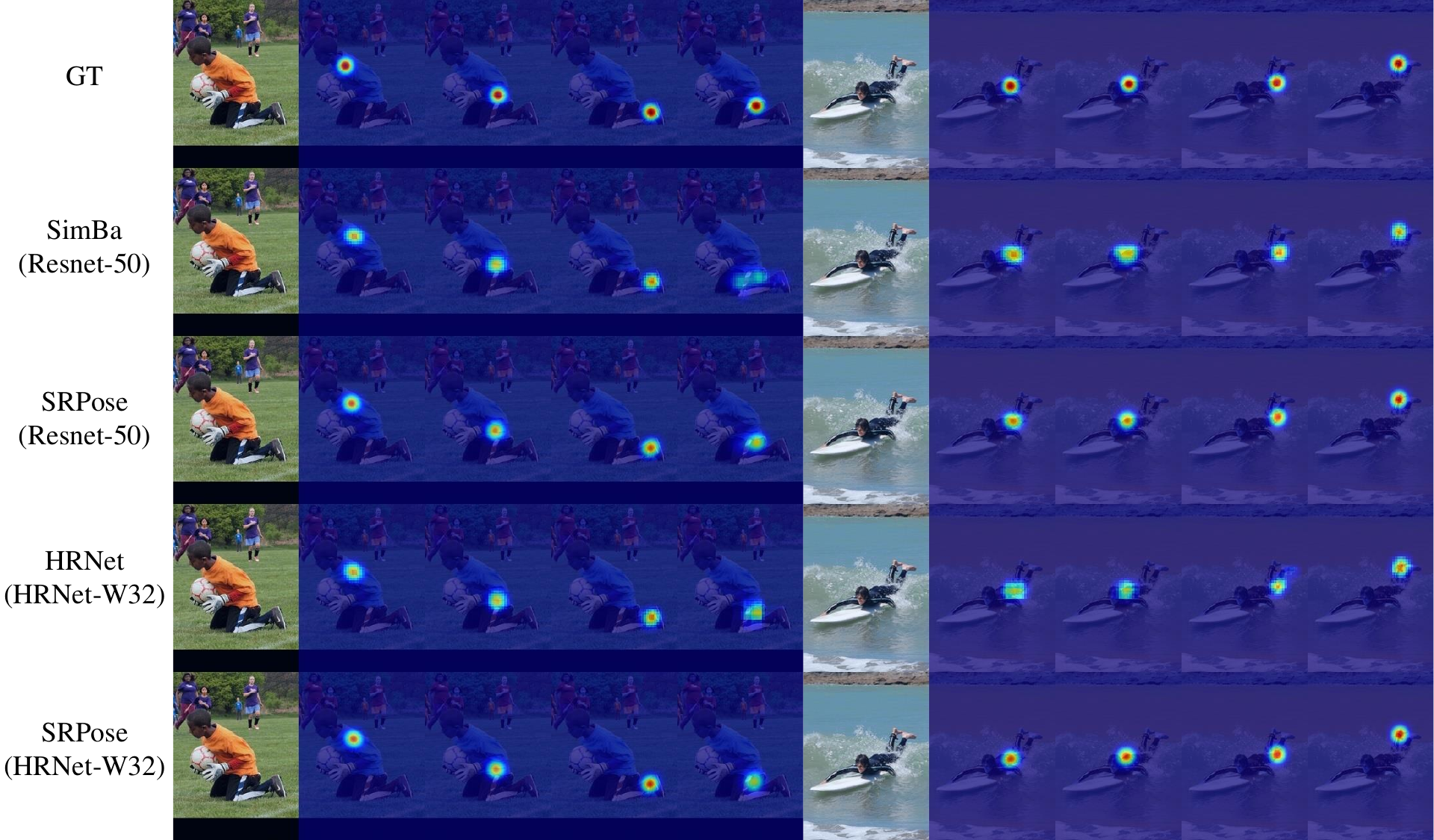}
    \end{center}
        \vspace{-0.1in}
    \caption{Heatmap visualization, where the first and sixth columns represent the input images, columns 2-5 represent the heatmaps of the images in column 2, and columns 7-10 represent the heatmaps of the images in column 2.}
    \label{fig:heatmap}
    \vspace{-0.2in}
\end{figure*}

To validate the effectiveness of our proposed SRPose, we compare the generated heatmaps with and without SRPose using HRNet and SimpleBaseline as baselines, which are shown in \cref{fig:heatmap}. Since the heatmaps generated by SRPose have a higher resolution, the visualization results show that the heatmaps generated by SRPose are clearer than the original baseline. Meanwhile, SRPose can help the model better recognize the keypoints and thus get more precise positions.

\newpage

\printbibliography